\documentclass{article}

\usepackage{microtype}
\usepackage{graphicx}
\usepackage{subcaption}
\usepackage{booktabs} 

\usepackage{csquotes}

\usepackage{hyperref}

\usepackage[accepted]{eiml_icml2026}

\usepackage{amsmath}
\usepackage{amssymb}
\usepackage{mathtools}
\usepackage{amsthm}

\usepackage[capitalize,noabbrev]{cleveref}

\theoremstyle{plain}

\newtheorem*{example*}{Example}

\theoremstyle{definition}

\theoremstyle{remark}

\usepackage[most]{tcolorbox}
\usepackage{xcolor}

\tcolorboxenvironment{example*}{
  enhanced,
  breakable,
  colback=orange!4,
  colframe=orange!45!black,
  boxrule=0.6pt,
  arc=2mm,
  left=2mm,
  right=2mm,
  top=1mm,
  bottom=1mm
}

\tcolorboxenvironment{remark}{
  enhanced,
  breakable,
  colback=gray!8,
  colframe=gray!50,
  boxrule=0.6pt,
  arc=2mm,
  left=2mm,
  right=2mm,
  top=1mm,
  bottom=1mm
}

\newtcolorbox{importantbox}{
  enhanced,
  breakable,
  colback=blue!3!white,
  colframe=blue!30!black,
  boxrule=0.5pt,
  arc=1.5mm,
  left=2mm,
  right=2mm,
  top=1.2mm,
  bottom=1.2mm
}

\newtcolorbox{takeawaybox}{
  enhanced,
  breakable,
  colback=red!3!white,
  colframe=red!30!black,
  boxrule=0.5pt,
  arc=1.5mm,
  left=2mm,
  right=2mm,
  top=1.2mm,
  bottom=1.2mm
}

\usepackage[textsize=tiny]{todonotes}

\newcommand{\mc}[1]{\mathcal{#1}}
\newcommand{\mb}[1]{\mathbb{#1}}
\newcommand{\td}[1]{\dot{#1}}

\newcommand{\maybeomit}[1]{}

\usepackage{tikz}

\newcommand{\circledlabel}[1]{%
  \tikz[baseline=(char.base)]{
    \node[draw,circle,inner sep=0.8pt] (char) {\footnotesize #1};
  }%
}

\definecolor{penStable}{RGB}{46, 139, 87}  
\definecolor{penUnstable}{RGB}{220, 70, 7}
\definecolor{penSaddle}{RGB}{198, 40, 40}
\definecolor{penSeparatrix}{RGB}{240, 180, 40}
\definecolor{penOscillation}{RGB}{25, 118, 210}
\definecolor{penRotation}{RGB}{200, 30, 30}

\definecolor{penIC0}{RGB}{25, 60, 150} 
\definecolor{penIC14}{RGB}{130, 180, 230} 

\definecolor{penIC}{RGB}{55, 71, 79} 
\definecolor{penDet}{RGB}{25, 118, 210}
\definecolor{penStoch}{RGB}{198, 40, 40}
\definecolor{penDensity}{RGB}{230, 120, 30}

\definecolor{penTrue}{RGB}{25, 118, 210}    
\definecolor{penObs}{RGB}{117, 117, 117}    
\definecolor{penFilt}{RGB}{230, 81, 0}     
\definecolor{penBand}{RGB}{255, 179, 0}     

\definecolor{penGammaLow}{RGB}{21, 101, 192}    
\definecolor{penGammaMid}{RGB}{123, 31, 162}   
\definecolor{penGammaHigh}{RGB}{198, 40, 40}  

\definecolor{penNL}{RGB}{25, 118, 210}   
\definecolor{penLin}{RGB}{220, 50, 50}   

\definecolor{revisionblue}{RGB}{0,80,160}

\icmltitlerunning{Submission and Formatting Instructions for ICML 2026}

\begin{document}

\twocolumn[
  \icmltitle{What Uncertainties Do We Need for Dynamical Systems?}



  \icmlsetsymbol{equal}{*}

  \begin{icmlauthorlist}
    \icmlauthor{Yusuf Sale}{1,3}
    \icmlauthor{Christopher Bülte}{2,3}
    \icmlauthor{Felix Czaja}{1}
    \icmlauthor{Joshua Stiller}{1,3}
    \icmlauthor{Eyke Hüllermeier}{1,3,4}
  \end{icmlauthorlist}

  \icmlaffiliation{1}{Institute of Computer Science, LMU Munich}
  \icmlaffiliation{2}{Department of Mathematics, LMU Munich}
  \icmlaffiliation{3}{Munich Center for Machine Learning (MCML)}
  \icmlaffiliation{4}{German Research Center for Artificial Intelligence (DFKI, DSA)}

  \icmlcorrespondingauthor{Yusuf Sale}{yusuf.sale@ifi.lmu.de}

  \icmlkeywords{Machine Learning, ICML}

  \vskip 0.3in
]



\printAffiliationsAndNotice{}  

\begin{abstract}
  The distinction between aleatoric and epistemic uncertainty has received considerable attention in machine learning research, mainly in the context of supervised learning but also in other settings such as generative modeling. 
  In this paper, we offer a machine learning perspective on uncertainty modeling for dynamical systems, which has been studied much less so far.
  In particular, we ask: what uncertainties do we need for dynamical systems? We discuss sources of uncertainty, clarify their nature (aleatoric or epistemic), and consider how the objectives of representing and quantifying uncertainty vary across different tasks. 
\end{abstract}

\section{Introduction} \label{sec:intro}
Uncertainty has become a central topic in machine learning (ML), with increasing interest in the distinction between aleatoric and epistemic uncertainty. Aleatoric uncertainty reflects randomness inherent in a process, whereas epistemic uncertainty originates from a lack of knowledge about that process. The former is therefore irreducible, while the latter can, in principle, be reduced by gathering more information \citep{hullermeier2021aleatoric}.

Increased interaction has been observed in recent years between ML and dynamical systems. For instance, neural differential equations parametrize dynamics through neural networks \citep{chen2018neural}, scientific ML \citep{RAISSI2019686,  karniadakis2021physics} couples learned components with governing equations, and data-driven control integrates learned models into control theory \citep{hewing2020learning}. This intersection has also made the aleatoric--epistemic dichotomy increasingly prominent in the dynamical systems setting \citep{psaros2023uncertainty}.

In this paper, we offer a perspective on the aleatoric–epistemic dichotomy in dynamical systems and ask: what uncertainties do we need? In this spirit, we revisit sources of uncertainty in dynamical systems, consider whether they are reducible or irreducible, and connect them to the ML literature. 
We further emphasize that the aleatoric–epistemic classification of a source depends on the task and modeling assumptions.
Finally, we discuss the two-way relationship between ML and dynamical systems, and identify opportunities for representing and quantifying uncertainty in dynamical systems.

\section{Dynamical Systems} \label{sec:ds}
Broadly speaking, a dynamical system is a rule that determines how a state evolves over time. Formally, let $\mc{X} \subseteq \mb{R}^d$ be a state space and $f: \mc{X} \rightarrow \mb{R}^d$ (in continuous time) be the evolution law. Given this law, the evolution of a continuous-time autonomous\footnote{A dynamical system is autonomous if the evolution law $f$ does not depend explicitly on time, i.e., the rule governing the system is the same at every time point $t$.} dynamical system is determined by its initial state. Or, stated differently, the evolution is given by the solution to the initial value problem 
\begin{equation}\label{eq:ds}
 \td{x}(t) = f(x(t)), \quad x(0) = x_0 \, , 
\end{equation}
where $x(t) \in \mc{X}$ is the state at time $t$ and $x_0 \in \mc{X}$ the initial condition. Under standard regularity conditions on $f$ (e.g., Lipschitz continuity), the Picard-Lindelöf theorem guarantees that \eqref{eq:ds} admits a unique solution $x(t) = \Phi_t(x_0)$ for all $t$ in some interval \citep{teschl2012ordinary}. The map $\Phi_t: \mc{X} \rightarrow \mc{X}$ is the flow of the system, satisfying $\Phi_0 = \rm{id}$ and the semigroup property $\Phi_{t + s} = \Phi_t \circ \Phi_s$. In this formulation, the system is a deterministic machine, i.e., given $x_0$, the future trajectory $\{x(t)\}_{t \geq 0}$ is uniquely determined. 

\begin{importantbox}
Throughout, we take the state $x(t)$ to be the quantity of interest, so when we ask what we are uncertain about, we mean \emph{the state of the system at time $t$}. We discuss this point in more detail in Section \ref{sec:need}. 
\end{importantbox}

Our focus is on finite-dimensional dynamical systems, that is, ODEs and, once process noise is introduced, SDEs. This setting is rich enough to illustrate the conceptual points we make while keeping the exposition concrete. The discussion extends naturally to partial differential equations, where the state is a function on a spatial domain, but some aspects become technically more intricate.

\begin{example*}[running]
Consider the nonlinear pendulum with viscous damping. The state $x = (\theta, \omega)^\top$\footnote{The natural state space of the pendulum is the phase cylinder $S^1 \times \mathbb{R}$. For notational simplicity, we use the lifted coordinate $\theta \in \mathbb{R}$ in equations and figures, and display finite covering intervals such as $[-2\pi,2\pi]$.} consists of the angular displacement $\theta$ and angular velocity $\omega = \dot{\theta}$, evolving according to 
\begin{equation} \label{eq:pend}
    \dot{\theta} = \omega, \quad \dot{\omega} = - \gamma \omega - \frac{g}{\ell} \sin \theta, 
\end{equation}
where $\gamma > 0$ is the damping coefficient and $g/\ell > 0$ is the ratio of gravitational acceleration to pendulum length. Throughout, we fix $\gamma = 0.5$ and $g/\ell = 1.0$ unless stated otherwise. 
\end{example*}

The phase portrait of this system (see Figure \ref{fig:1}, \autoref{app:run}) exhibits a rich structure (stable and unstable equilibria, separatrices, qualitatively distinct dynamical regimes) that will play a role in how uncertainty propagates in later sections.
At this stage, there is no uncertainty. Every component of the system, namely the state space, the parameters $(\gamma, g/\ell)$, the function $f$, and the initial condition $x_0$, is known exactly, so the state $x(t)$ is determined for every $t$. 

\section{Sources of Uncertainty}
In practice, however, the system is rarely fully determined. The question is: where does uncertainty enter, and what is its nature? In the ML literature, a distinction between \emph{types} of uncertainty and \emph{sources} of uncertainty has been made \citep{hullermeier2021aleatoric, gruber2023sources}. The type of uncertainty refers to its nature, namely whether it is aleatoric (irreducible) or epistemic (reducible). The source of uncertainty, on the other hand, answers a different question: not what kind of uncertainty we face, but \emph{why} it arises. We adopt a similar viewpoint for dynamical systems. Starting from the deterministic baseline \eqref{eq:ds}, we ask: which components of the system may be uncertain, and how does this uncertainty affect our knowledge of $x(t)$? Each component defines a distinct source of uncertainty. For each source, we may further ask whether the resulting uncertainty is aleatoric or epistemic.

\circledlabel{S1} \textbf{Initial condition uncertainty.} The deterministic baseline assumes that the initial state $x_0$ is known exactly. However, this is not always the case, since initial states are obtained from measurements, estimates, or prior knowledge.  
Formally, a natural way to represent this uncertainty is to replace the point-valued initial condition $x_0 \in \mc{X}$ with a probability distribution $p_0$ over the state space: $x(0) \sim p_0(x)$.
Instead of a trajectory $\Phi_t(x_0)$, we must now consider the image of the initial distribution under the flow. Thus, each possible initial state $x_0$ gives rise to its own trajectory $\Phi_t(x_0)$, and the distribution over initial states induces, at every time $t$, a distribution over current states. Technically, the evolution of the distribution is governed by the Liouville equation, 
\begin{equation} \label{eq:lv}
    \frac{\partial p_t}{\partial t}(x) = - \nabla \cdot (f(x) p_t(x)), 
\end{equation}
a partial differential equation (PDE) describing the transport of probability mass by the vector field $f$ \citep{teschl2012ordinary}. 
\maybeomit{A central insight is that the flow not only propagates uncertainty, it reshapes it. The local rate of deformation is governed by the Jacobian $\nabla f(x)$; directions along which the flow expands amplify uncertainty, while directions along which it contracts suppress it. For linear systems $\dot{x} = Ax$, this is captured by the eigenvalues of $A$; an initial Gaussian $p_0 = \mc{N}(\mu_0, \Sigma_0)$ remains Gaussian at all times, with mean $\mu_t = e^{At}\mu_0$ and covariance $\Sigma_t = e^{At}\Sigma_0 e^{A^{\top}t}$, which grows exponentially along unstable eigendirections and shrinks along stable ones. For nonlinear systems, the deformation is state-dependent and can be far more severe.}

This dependence on phase-space geometry is what distinguishes uncertainty in dynamical systems from uncertainty in static models. In a dynamical system, the transformation acts continuously, and the amplification or suppression of uncertainty depends on the local structure of the flow at the current state, which changes as the state evolves. Uncertainty in dynamical systems is not static, it is a process.

\circledlabel{S2} \textbf{Process noise.} 
We now consider a qualitatively different source, randomness in the dynamics itself. The ODE \eqref{eq:ds} is replaced by a stochastic differential equation (SDE),  
\begin{equation} \label{eq:sde}
 dx(t) = f(x(t)) dt + \sigma(x(t)) dW_t, \quad x(0) = x_0,    
\end{equation}
where $W_t$ is an $m$-dimensional standard Wiener process and $\sigma: \mc{X} \rightarrow \mb{R}^{d \times m}$ is the diffusion coefficient, controlling the intensity and directional structure of the noise. The system is no longer a deterministic machine; thus, even with perfect knowledge of the initial condition $x_0$, the trajectory $\{x(t)\}_{t\geq 0}$ is a stochastic process, and repeating the experiment from the same starting point will produce a different realization each time. Earlier, uncertainty was \enquote{injected} once at $t=0$, and subsequently transformed by the flow. Here, uncertainty is injected continuously; at every instant, the noise term $\sigma dW_t$ perturbs the state. Thus, uncertainty cannot be reduced through (time) evolution alone. In a dissipative\footnote{A dissipative system is one that loses energy over time.} deterministic system, an initial cloud of states contracts onto the attractor and uncertainty shrinks. In the stochastic system \eqref{eq:sde}, the attractor still pulls states inward, but the noise continuously replenishes the uncertainty that dissipation resolves. The result is a dynamic equilibrium between contraction and diffusion. This equilibrium is made precise by the Fokker-Planck equation, which governs evolution of the probability density $p_t(x)$ under the SDE \eqref{eq:sde}: 
\begin{equation}\label{eq:fpe}
\frac{\partial p_t}{\partial t}(x) = - \nabla \cdot \left( f(x)p_t(x) \right) + \frac{1}{2} \sum_{i,j} \frac{\partial^2}{\partial x_i \partial x_j}\left[D_{ij}(x) p_t(x) \right],
\end{equation}
where $D(x) = \sigma(x)\sigma(x)^{\top}$ is the diffusion tensor \citep{risken1984fpe}. The first term on the right-hand side is identical to the Liouville equation \eqref{eq:lv} and describes transport of probability by the vector field $f$. The second term is a diffusion term that spreads probability, even in regions where the flow is zero. The Liouville equation is the $\sigma \rightarrow 0$ limit of \eqref{eq:fpe}. 

\maybeomit{For dissipative systems with sufficient noise, \eqref{eq:fpe} admits a stationary solution $p_{\infty}(x)$ satisfying $\partial p_{\infty} / \partial t = 0$, a time-independent density representing the long-run balance between contracting flow and dispersing noise. The stationary distribution has no analog in the earlier initial condition setting. There, if the system is dissipative, the density collapses onto the attractor and uncertainty vanishes asymptotically. Here, $p_{\infty}$ has non-zero spread, which is determined by the ratio of noise intensity to dissipation strength (i.e., stronger noise or weaker damping yields a broader stationary distribution). The existence of $p_{\infty}$ implies that the system forgets its initial condition. Regardless of where the system starts, $p_t \rightarrow p_{\infty}$ as $t \rightarrow \infty$. After sufficient time, the initial condition uncertainty becomes irrelevant, and all remaining uncertainty is due to process noise. The timescale on which this forgetting occurs (\emph{viz.} the mixing time) is a fundamental property of the system, determined by the spectral gap of the Fokker-Planck operator. }

\circledlabel{S3} \textbf{Partial observations.} 
We have implicitly assumed that the full state is directly observable throughout. In practice, however, this is rarely the case. Physical systems are observed through sensors that may measure only some components of the state, and introduce their own noise. Formally, we augment the dynamical system with an observation equation. At discrete times $t_1, t_2, \dots$, we receive measurements 
\begin{equation*}
    y_k = h(x(t_k)) + v_k, \quad v_k \sim \mc{R},
\end{equation*}
where $h: \mc{X} \rightarrow \mb{R}^q$ is the observation function, which maps the full state to the quantity we actually measure, and $v_k$ is observation noise, drawn independently at each measurement time from some distribution $\mc{R}$. 
The dimension $q$ of the observation may be smaller than the state dimension $d$.

Reconstructing the state $x(t_k)$ from observations $y_1,\dots,y_k$ is the filtering problem \citep{jazwinski2007stochastic}. In the Bayesian formulation, the goal is to maintain a posterior distribution $p(x(t_k)\,|\, y_{1:k})$ that combines what the dynamics predict with what the observations reveal \citep{sarkka2023bayesian}.  
Filtering follows a predict-update rhythm. 
Between observations, uncertainty is propagated by the dynamics and may grow, especially in the presence of process noise or unstable directions. Stable dynamics may instead contract it. Observations typically reduce uncertainty in the observed directions.
For linear-Gaussian systems, the Kalman filter realizes this cycle \citep{kalman1960new}, maintaining a Gaussian posterior $\mc{N}(\hat{x}_k, P_k)$ whose covariance $P_k$ grows during prediction and shrinks during updates. For nonlinear/non-Gaussian systems, approximate methods (extended Kalman filter \citep{jazwinski2007stochastic}, particle filters \citep{gordon1993novel}, etc.) are required, but the structure remains the same.  

\circledlabel{S4} \textbf{Parameter uncertainty.}
We now consider the possibility that the dynamics themselves are not fully known, because they depend on parameters whose values are uncertain. Formally, the evolution law is parametrized as $f_{\theta}$, where $\theta \in \Theta$ is a vector of parameters. The system 
\begin{equation} \label{eq:param}
    \dot{x}(t) = f_{\theta}(x(t)) 
\end{equation}
is fully specified only when $\theta$ is known. If $\theta$ is uncertain, then the trajectory is uncertain even if the initial condition, observation process and model structure are all known.

The task of estimating $\theta$ from observed trajectories is the system identification problem \citep{lennart1999system}, and in the Bayesian formulation it amounts to computing the posterior $p(\theta \, | \, \mc{D})$. What distinguishes parameter uncertainty from other sources discussed so far is how it propagates through time. Initial condition uncertainty \circledlabel{S1} is injected once and then transformed by the flow. Process noise \circledlabel{S2} is injected continuously but independently at each instant. Parameter uncertainty is different from both because $\theta$ is fixed but unknown and influences the dynamics at every instant in the same way. The trajectory $x(t)$ depends on $\theta$ through the entire history of the dynamics, since $x(T) = \Phi^{\theta}_{T}(x_0)$, where $\Phi^{\theta}_{T}$ denotes the flow map under the parametrized dynamics. A small error in $\theta$ does not produce a single perturbation, it produces a systematic bias that compounds over time. 
The sensitivity $\partial x(T)/\partial \theta$ is governed by accumulated linearizations along the trajectory. In discrete time these appear as products of Jacobians, whereas in continuous time they are described by the variational equation and its associated state-transition operator. In unstable or chaotic regimes, these accumulated linearizations can grow exponentially with the time horizon.
This compounding can have qualitative consequences. Near a bifurcation boundary, the posterior $p(\theta \,|\, \mc{D})$ may leave us uncertain not about how much but about what kind of behavior the system exhibits, for instance, whether it has a stable equilibrium, a limit cycle, or chaotic dynamics. The uncertainty becomes qualitative rather than quantitative.

\circledlabel{S5} \textbf{Structural uncertainty.} 
The sources of uncertainty considered so far all presuppose that the form of the dynamical system is correct. We now relax this assumption and ask what happens when the model itself is wrong. Structural uncertainty encompasses several distinct possibilities: 

\emph{(i)} \emph{Wrong functional form.} The vector field $f$ may be specified within an inappropriate function class. For example, one may assume linear dynamics although the truth is nonlinear, use a polynomial expansion although the true dependence is transcendental, or choose a neural network architecture that is incapable of representing the true vector field.

\emph{(ii)} \emph{Missing state variable.} The state space $\mc{X}$ might be too small. The system may possess hidden degrees of freedom that influence the observed dynamics without being represented in the model. What appears to be a low-dimensional system is in fact a projection of a higher-dimensional one.

\emph{(iii)} \emph{Wrong type of dynamics.} We may model the system as autonomous although external forcing is present, as an ODE although the dynamics involve delays or memory effects, or as deterministic although intrinsic stochasticity is essential.

\emph{(iv)} \emph{Wrong observation model.} The observation function $h$ may be misspecified, or the noise distribution $\mc{R}$ may not reflect reality.

\subsection{Interactions and Temporal Behavior}
In real dynamical systems, several sources usually act simultaneously, and their interactions are non-trivial. We discuss how these sources compound and how their relative importance shifts over the prediction horizon.

\emph{Compounding.} Consider a system with both uncertain initial conditions \circledlabel{S1} and uncertain parameters \circledlabel{S4}. Initial condition uncertainty alone is transported by a known flow $\Phi_t$, with its deformation governed by the Jacobian of $f$. Parameter uncertainty, by contrast, propagates through the dynamics via the sensitivity $\partial x(t) / \partial \theta$, which itself accumulates over time. When both are present, these effects no longer separate cleanly, because the flow acting on the initial cloud itself depends on the uncertain parameter. Different parameter values generate different dynamics and therefore deform the same initial cloud in different ways. 

\emph{Horizon-dependent dominance.} Different sources dominate at different prediction horizons, and their relative importance shifts in a characteristic way. Over short horizons, initial condition uncertainty \circledlabel{S1} often dominates, because the system has not had time to forget where it started, and perturbations have not yet been strongly amplified or suppressed by the flow. Over intermediate horizons, process noise \circledlabel{S2} accumulates, and observation-induced uncertainty \circledlabel{S3} becomes increasingly relevant, especially along unstable directions. Over long horizons, parameter uncertainty \circledlabel{S4} and structural uncertainty \circledlabel{S5} tend to dominate, because they govern the long-run dynamics and, in particular, the attractor towards which the system evolves. On sufficiently long timescales, trajectories lie near their attractor largely regardless of their starting point or the particular noise realizations encountered along the way. What differs across the ensemble is then not the transient evolution, but the attractor itself, as determined by $\theta$ and the form of $f$.

\section{Nature of Uncertainty}\label{sec:need}

Process noise  \circledlabel{S2} is the prototypical example of aleatoric uncertainty, namely irreducible randomness inherent to the dynamics. No  additional data can eliminate it, because it arises from the evolution law itself. The noise intensity $\sigma$, however, is typically not known from first principles and must be estimated from data. 
What is specific to the dynamical setting is that $\sigma$ enters continuously, at every instant, so errors in $\sigma$ compound along the trajectory and affect the stationary distribution, the mixing time, and probabilities of rare noise-driven transitions in ways that have no static analog.

Other uncertainties appear to be mainly of epistemic nature. One can argue, for example, that the system has a true initial state $x_0$, even if we do not know it exactly. Then, the distribution $p_0$ represents ignorance that could, in principle, be reduced by acquiring more information (e.g., more precise measurements). Likewise, partial observations \circledlabel{S3} are a source of epistemic uncertainty: the system has a definite state $x(t_k)$, and we are simply ignorant of it, but the extent to which this uncertainty is reducible has structural limits. In observable systems, the posterior uncertainty over the state converges to a steady level determined by the noise. In unobservable systems, there exist directions in state space that no amount of observations can distinguish, and epistemic uncertainty along those directions persists indefinitely, regardless of measurement effort.

It is natural to ask how this relates to uncertainty about the initial state. The two sources are distinct, yet connected. In practice, uncertain initial conditions often arise from partial observations, since the initial state was estimated from the same imperfect sensors that will provide measurements going forward. Conversely, the filtering problem exists even when the initial state is known exactly.

Parameter uncertainty \circledlabel{S4} is epistemic, too: the system has a true parameter (e.g., gravity does not fluctuate between experiments) and our uncertainty reflects a lack of knowledge that is, in principle, reducible by collecting more data. The same applies to structural uncertainty, which, however, is qualitatively distinct from other sources discussed so far. Parameter uncertainty is reducible within a model class: given enough data, the posterior $p(\theta \,|\, \mc{D})$ concentrates on the truth (if the truth is in the class). Observation-induced uncertainty \circledlabel{S3} is reducible within the limits of observability. In both cases, the relevant notion of \enquote{the truth} is well-defined relative to the model. Structural uncertainty, by contrast, is uncertainty about \enquote{the right model,} and this question cannot be answered from inside any single model, nor with additional data.  
This echoes the classical notion of model misspecification in statistics and ML, where the true data-generating process lies outside the hypothesis class considered.
This makes structural uncertainty notoriously difficult to represent (and quantify). The \enquote{unknown unknowns} are, almost by definition, beyond the reach of any procedure operating within a fixed hypothesis space.

In some cases, the divide between aleatoric and epistemic uncertainty is not entirely obvious, and also depends on the assumptions and modeling choices we make. Consider again the initial condition. As mentioned before, if we model $x_0$ as a fixed but unknown quantity inferred from imperfect measurements, the uncertainty is epistemic and can in principle be reduced with a better sensor, provided such a sensor is available. If, however, we treat $x_0$ as a random variable drawn from a distribution that represents natural variability in the setup, for instance when a human releases a pendulum from a small displacement, with a slightly different release each time, the uncertainty is aleatoric. The same point has been made in ML \citep{hullermeier2021aleatoric}: A source of uncertainty is not categorized as aleatoric or epistemic per se; instead, the categorization depends on the assumptions and modeling setup.

The need or usefulness of disentangling aleatoric and epistemic uncertainty depends on the task at hand. Indeed, dynamical systems are studied through a range of tasks, each with its own object of interest. 
Forecasting \citep{gneiting2014probabilistic}, filtering and smoothing \citep{sarkka2023bayesian} all take the state itself as their object of interest, differing only in the time at which it is queried and in the observations on which the inference is based. The object of interest is, respectively, the future state $x(t + \Delta t)$ given the present, the present state $x(t)$ given past and present observations, or a past state $x(s)$ given the full observation window. 
System identification \citep{lennart1999system} shifts the interest from the state to the parameters $\theta$ of $f_{\theta}$, or even the form of $f$ itself; trajectories then only serve as evidence from which the model is inferred. 
Verification \citep{mitchell2005time, bansal2017hamilton} translates task-specific uncertainties into a reachable set, 
whereas control \citep{aastrom2021feedback} uses them to guide the choice of an input $u(t)$. Verification and control are examples of downstream tasks: rather than representing uncertainty about an object of interest (e.g., the state $x(t)$) for its own sake, they leverage uncertainty to support a separate goal, e.g., a guarantee in verification or a control decision. Here, invoking the aleatoric-epistemic dichotomy becomes operationally consequential, and in some cases essential. Control is the clearest example. Epistemic uncertainty has a value of information and can be reduced by acting on the system, while aleatoric uncertainty cannot, and the two therefore demand fundamentally different actions.

\begin{takeawaybox}
    Whenever we speak about uncertainty in a dynamical system, we are implicitly committing to a task, to an object of interest, and to a modeling setup. Making the commitments explicit is a prerequisite for categorizing uncertainty and judging the usefulness of disentangling aleatoric and epistemic uncertainty.
\end{takeawaybox}

\section{How to Model the Uncertainties?} \label{sec:model}

As in many other scientific disciplines, the prevailing paradigm for modeling uncertainty in dynamical systems is probability theory. That is, uncertainty is mainly perceived as a form of randomness. An obvious example is stochastic differential equations, which, as explained before, model random influences on the evolution of a dynamical system using a Wiener process as the source of stochasticity. 

With regard to the aleatoric/epistemic divide, it has been argued that standard probability distributions are insufficient for representation, because they solely capture aleatoric but no epistemic uncertainty \citep{sale2026aleatoric}. Therefore, considering epistemic uncertainty as uncertainty about the ground-truth distribution, second-order distributions has been proposed \cite{hullermeier2021aleatoric}. By now, modeling epistemic uncertainty in terms of second-order distributions is well established in standard ML. 

Could this approach also be used for dynamical systems? In principle, this is possible, although a second-order evolution of a dynamical system will be challenging from both representational and computational viewpoints. Perhaps more importantly, whether a second-order distribution is meaningful depends on assumptions about the ground truth. In supervised learning, a second-order distribution is required to represent epistemic uncertainty, since the truth that the learner is attempting to predict is, by assumption, a distribution (namely, the conditional distribution of outcomes $y$ given an input $x$). In dynamical systems, this is not necessarily the case. For example, one may plausibly assume that the initial state $x_0$ is a precise albeit ill-known quantity. Then, a first-order distribution (interpreted subjectively) is enough to represent epistemic uncertainty about this quantity. 

A simple alternative to probability distributions is uncertainty representation in terms of \emph{sets}. For example, an imprecisely known initial state $x_0$ might be represented by a region $X_0 \subseteq \mathcal{X}$ with guaranteed coverage. Interestingly, \emph{differential inclusions} make use of sets to model uncertainty about the right-hand side of a differential equation: an equation $\dot{x} = f(x)$ is replaced by an inclusion $\dot{x} \in F(x)$, where $F$ is a set-valued function $\mathcal{X} \rightarrow 2^{\mathbb{R}^d}$. Correspondingly, system states $x(t) \in \mathcal{X}$ are transformed into reachable sets $X(t) \subset \mathcal{X}$. There is an extensive literature on fuzzy differential equations (inclusions), where sets are replaced by more general fuzzy sets \cite{fuzzyde}.

Sets can also serve as a second-order representation: a set of probability distributions is referred to as a \emph{credal set} \citep{wall_sr}. Credal machine learning, i.e., training models that produce credal sets rather than second-order distributions as predictions, has recently attracted attention \cite{wangCredalWrapper2025,loehrCredalPrediction2025}. From an uncertainty representation point of view, sets seem less informative than distributions. Yet, set-based representations also have advantages. In particular, sets are often easier to specify. Moreover, many have argued that probability distributions are less suitable for representing epistemic uncertainty, i.e., ignorance in the sense of a lack of knowledge \citep{dubo_rp96}. 

\section{Machine Learning and Dynamical Systems} \label{sec:ml-ds}

In recent years, the research fields of dynamical systems and ML have started to converge. We reflect briefly on what each field contributes to the other, putting mutual benefits in the context of our earlier discussion.

\subsection{Machine Learning for Dynamical Systems}
The most visible contribution of ML to dynamical systems has been the expansion of the class of models one can practically work with. 
Classical system identification \citep{lennart1999system} proceeds from parametric model classes and fits parameters to data. ML offers flexible, high-capacity function approximators that can represent dynamics too complex for traditional parametric models.
Neural ordinary differential equations \citep{chen2018neural} parametrize the right-hand side of an ODE with a neural network and learn it end-to-end from trajectories.  
Neural stochastic differential equations \citep{tzen2019neural, li2020scalable, kidger2021neural, kidger2021efficient} extend this idea to genuinely stochastic systems, with both drift and diffusion parametrized by neural networks. In this way, they address uncertainty from process noise \circledlabel{S2} directly at the modeling level.
Gaussian-process state-space models \citep{frigola2013bayesian, frigola2014variational} follow a nonparametric approach and place a GP prior over the transition function, which yields calibrated posteriors over the dynamics. 
This addresses the functional-form component of structural uncertainty \circledlabel{S5} within a chosen state-space and kernel framework.
Sparse regression methods such as SINDy \citep{brunton2016discovering, brunton2016sparse} represent the opposite end of the modeling spectrum. They seek interpretable and parsimonious governing equations from data. This is useful in scientific applications, where the inferred structure provides valuable insights and domain knowledge (cf.\ Section \ref{sec:need}).

These model classes are highly expressive but do not necessarily ensure reliable uncertainty representation. A parallel line of work has therefore focused on principled probabilistic treatments. Bayesian neural ODEs \citep{dandekar2020bayesian} place priors over the weights of neural ODEs and use MCMC or variational inference to obtain posteriors. Thus, they target parameter uncertainty \circledlabel{S4} in a data-driven setting where the parametric form is itself given by a neural network. Deep Kalman filters and related models \citep{krishnan2015deep, krishnan2017structured} combine classical filtering with neural function approximators. They support inference in state-space models with transition or emission distributions that exceed the scope of linear-Gaussian machinery and therefore bear directly on the partial-observation source \circledlabel{S3}.

A parallel literature extends these ideas to PDEs, where the state is a function rather than a finite vector. Neural operators \citep{JMLR:v24:21-1524}, DeepONets \citep{Lu_2021} and physics-informed neural networks (PINNs) \citep{RAISSI2019686} have succeeded in modeling PDE-based dynamical systems, with PINNs encoding known equations as soft training constraints and operator-learning approaches learning solution maps directly. Several extensions have been proposed and differ mainly in how they represent uncertainty. Input-perturbation approaches model uncertain initial conditions stochastically and propagate the resulting uncertainty through the model to obtain a probabilistic output \citep{pathak2022fourcastnetglobaldatadrivenhighresolution, biAccurateMediumrangeGlobal2023}. Post-hoc approaches fit a probability distribution on top of an already trained model \citep{magnani2025linearization, bulte_uncertainty_2025}. Fully generative methods replace the deterministic model with a probabilistic one that directly outputs samples or distributions over trajectories \citep{kohl_benchmarking_2024, price_probabilistic_2025}.

\subsection{Dynamical Systems for Machine Learning}
The converse direction is less often phrased explicitly but is equally substantive. Bayesian filtering and smoothing \citep{kalman1960new, sarkka2023bayesian} offer ML a principled framework for sequential inference under partial observability. This influence is visible both in specific architectures\,---\,deep state-space models for probabilistic time-series forecasting \citep{rangapuram2018deep}, structured state-space layers for long-range sequence modeling \citep{gu2021efficiently}\,---\,and in broader methodological ideas, such as model-based reinforcement learning with learned probabilistic dynamics \citep{deisenroth2011pilco}.

Control theory \citep{aastrom2021feedback} offers a language for reasoning about uncertainty and action. The exploration-exploitation tradeoff in modern reinforcement learning \citep{sutton1998reinforcement} is, at root, the insight that aleatoric and epistemic uncertainty call for different actions. Reachability analysis \citep{mitchell2005time, bansal2017hamilton} provides worst-case, set-valued guarantees that have become central to safe reinforcement learning and verified ML.

Dynamical systems have also shaped the understanding of ML architectures and their training. Deep residual networks can be interpreted as discretized ODEs \citep{Haber_2017}, an insight that directly motivated the neural ODE architecture \citep{chen2018neural}. Optimal control recasts neural network training as a control problem \citep{E_2018, JMLR:v18:17-653, BONNET2023113161} and offers tools for stability and robustness analysis  \citep{miller2019stablerecurrentmodels, 10179161, 10.5555/3454287.3455312}.

\section{Opportunities} \label{sec:out}
The discussion above suggests several opportunities for future work, where recent advances in ML offer natural starting points. Transferring second-order methods from supervised learning to dynamical systems is one such direction. The adaptation is non-trivial, however, since a second-order distribution over trajectories is more complex than one over a single prediction. A second, more distinctive question concerns what uncertainty quantification should measure in a dynamical system. In supervised learning, uncertainty is typically quantified at the level of a single prediction; in a dynamical system, the natural object is a trajectory, or a functional of one. This opens room for new measures. Uncertainty in derived quantities, such as the time to stabilization or the probability of entering a given region, may be more useful operationally than uncertainty in the state itself. Each direction invites us to rethink what uncertainty means when the object of interest is no longer static.

\section*{Acknowledgements}
YS and CB are supported by the DAAD program Konrad Zuse
Schools of Excellence in Artificial Intelligence, sponsored
by the Federal Ministry of Education and Research. FC gratefully acknowledges funding by the German Research Foundation (Deutsche Forschungsgemeinschaft, DFG) – GRK 3081 – Project number 534429653.

\bibliography{references}
\bibliographystyle{apalike}

\newpage
\appendix
\onecolumn

\newpage

\section{Running Example}\label{app:run}
\subsection*{Physical setup and phase portrait}
\begin{figure}[!h]
    \centering
    \includegraphics[width=\linewidth]{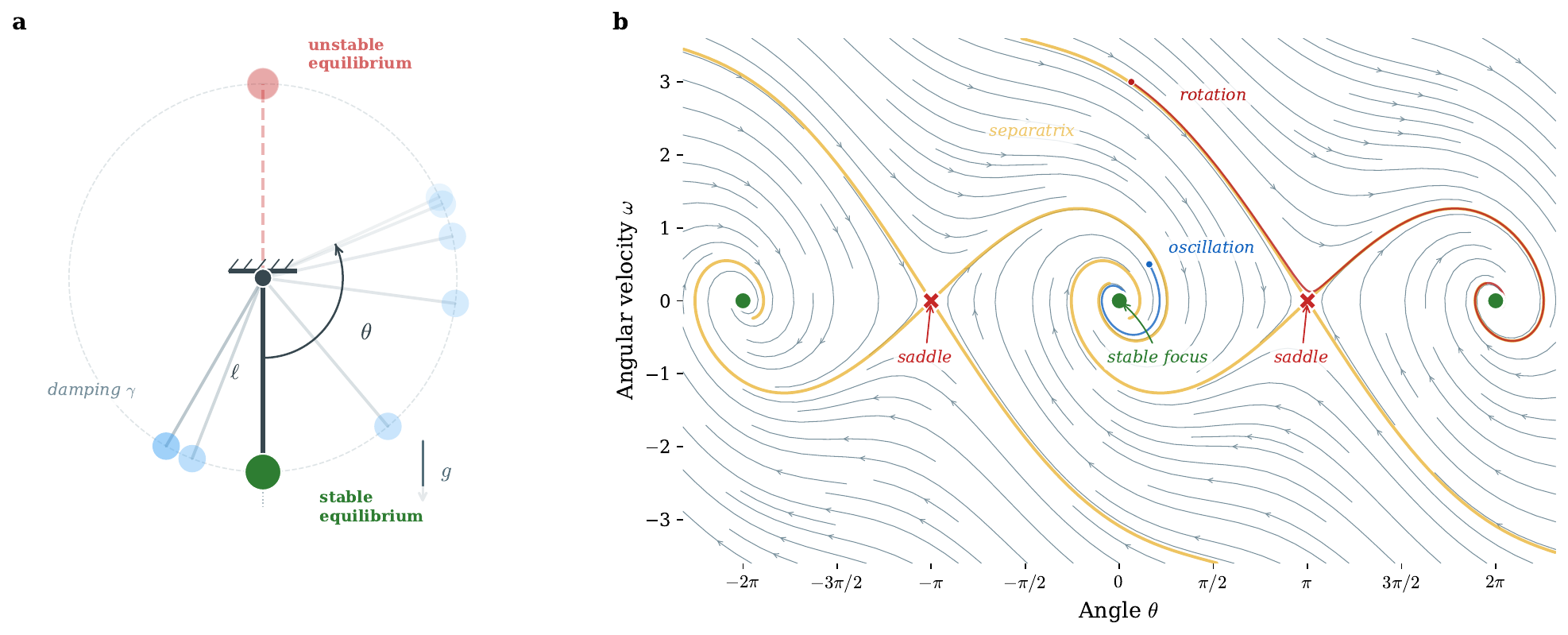}
    \caption{\textbf{(a)} Physical setup. A pendulum of length $\ell$ and angle $\theta$ swings under gravity $g$ with damping $\gamma$. It has a \textcolor{penStable}{stable equilibrium} (\textcolor{penStable}{$\bullet$}) at the bottom and an \textcolor{penUnstable}{unstable equilibrium} (\textcolor{penUnstable}{$\bullet$}) at the top (i.e., the inverted position). \textbf{(b)} Phase portrait of the system in the $(\theta, \omega)$-plane. The dynamics exhibit a \textcolor{penStable}{stable focus} (\textcolor{penStable}{$\bullet$}) at the origin, two \textcolor{penSaddle}{saddle points} (\textcolor{penSaddle}{$\times$}) at $(\pm\pi, 0)$, and \textcolor{penSeparatrix}{separatrices} (\textcolor{penSeparatrix}{---}) that divide the phase plane into qualitatively distinct regions. Trajectories inside the separatrices \textcolor{penOscillation}{oscillate} (\textcolor{penOscillation}{---}) and spiral toward rest, while trajectories outside them correspond to full \textcolor{penRotation}{rotations} (\textcolor{penRotation}{---}) that gradually slow. The angle axis is shown over $[-2\pi, 2\pi]$ to emphasize the periodic structure. }
    \label{fig:1}
\end{figure}
\clearpage

\begin{figure}[!h]
    \centering
    \includegraphics[width=\linewidth]{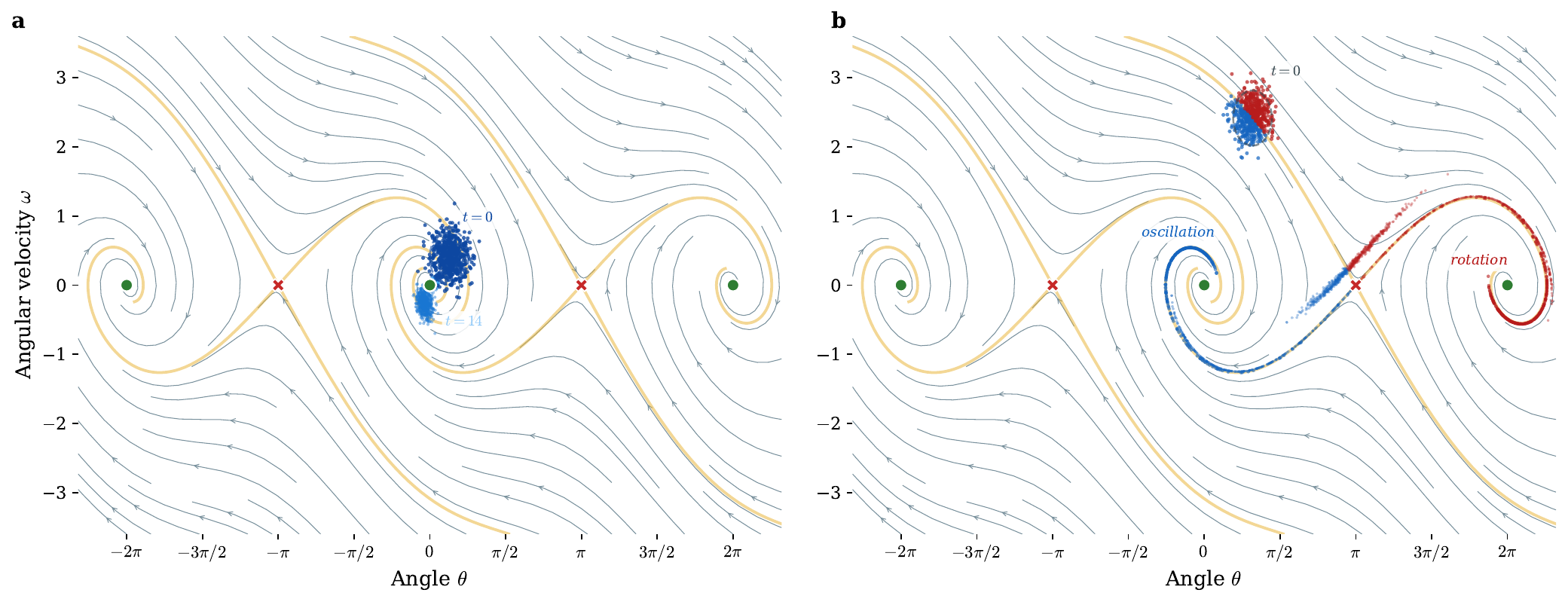}
    \caption{An ensemble of $1000$ initial states is drawn from a Gaussian at $t=0$ and evolved under the deterministic dynamics. \textbf{(a)} A cloud that stays well inside a \textcolor{penSeparatrix}{separatrix} (\textcolor{penSeparatrix}{---}) contracts as trajectories spiral toward the \textcolor{penStable}{stable focus} (\textcolor{penStable}{$\bullet$}), and uncertainty about $x(t)$ shrinks with time, compare $t=0$ (\textcolor{penIC0}{$\bullet$}) with $t=14$ (\textcolor{penIC14}{$\bullet$}). \textbf{(b)} A cloud straddling a separatrix is torn apart: part of the ensemble enters the \textcolor{penOscillation}{oscillation} (\textcolor{penOscillation}{$\bullet$}) regime and spirals toward the origin, while the rest enters the \textcolor{penRotation}{rotation} (\textcolor{penRotation}{$\bullet$}) regime and circulates around the cylinder.}
    \label{fig:2}
\end{figure}

\circledlabel{S1} \textbf{Initial condition uncertainty.}
Consider an initial distribution $p_0$, for example a Gaussian cloud concentrated around $(\theta_0,\omega_0)$, in the phase plane of the pendulum \eqref{eq:pend}. If the bulk of this cloud lies near the origin, the flow contracts it over time, i.e., all trajectories spiral inward, and the initial uncertainty shrinks as the ensemble converges towards rest (Figure \ref{fig:2}, \textbf{a}). The situation is qualitatively different when the cloud straddles the separatrix. Trajectories starting just inside the separatrix swing back and oscillate towards the origin; trajectories starting just outside go over the top before settling. The ensemble is \enquote{torn} apart by the flow, i.e., a single localized initial distribution gives rise to a bimodal distribution over qualitatively distinct dynamical behaviors (Figure \ref{fig:2}, \textbf{b}). No amount of temporal evolution will bring these two groups back together, the separation is irreversible and grows before it saturates. Thus, the geometry of the phase portrait determines how initial condition uncertainty is transformed by the dynamics. Regions of high dynamical sensitivity, such as neighborhoods of separatrices and saddle points, act as amplifiers of uncertainty, while regions of strong stability act as suppressors.
\clearpage

\begin{figure*}[!h]
    \centering
    \includegraphics[width=\linewidth]{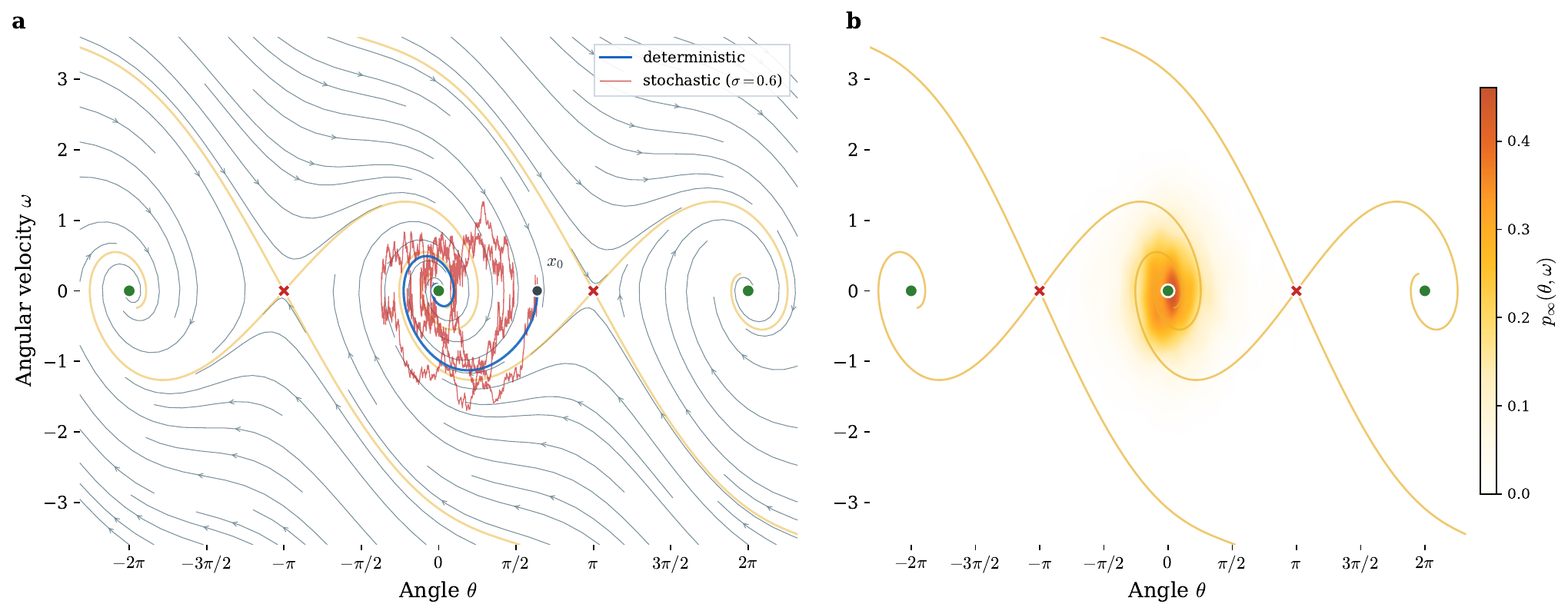}
    \caption{\textbf{(a)} Two trajectories from the same initial condition $x_0$ (\textcolor{penIC}{$\bullet$}) under the same dynamics: the \textcolor{penDet}{deterministic} (\textcolor{penDet}{---}) flow spirals smoothly toward the \textcolor{penStable}{stable focus} (\textcolor{penStable}{$\bullet$}), while the \textcolor{penStoch}{stochastic} (\textcolor{penStoch}{---}) trajectory with $\sigma=0.6$ jitters continuously and never settles to a point. \textbf{(b)} The stationary distribution $p_\infty(\theta, \omega)$ (\textcolor{penDensity}{\rule[0.5ex]{1.2em}{1pt}}) obtained from a long stochastic simulation.}
    \label{fig:3}
\end{figure*}

\circledlabel{S2} \textbf{Process noise.} Consider the pendulum \eqref{eq:pend} subject to additive noise in the angular velocity, 
\begin{equation} \label{eq:pn}
    d\theta = \omega dt, \quad d\omega = \left(- \gamma \omega - \frac{g}{\ell} \sin \theta \right) dt + \sigma dW_t.
\end{equation}
Physically, this represents random torques (e.g., vibrations of the pivot, fluctuating air currents, \ldots) that continuously perturb the pendulum's motion. Unlike the deterministic system, the pendulum never comes to rest. It fluctuates around the equilibrium, occasionally swinging to larger amplitudes when a sequence of kicks adds up coherently (Figure \ref{fig:3}, \textbf{a}). The stationary distribution $p_{\infty}(\theta, \omega)$ concentrates near the origin but has tails extending outward, reflecting a balance between the deterministic attractor pulling inward and the noise pushing outward (Figure \ref{fig:3}, \textbf{b}).
\clearpage 

\begin{figure*}[!h]
    \centering
    \includegraphics[width=\linewidth]{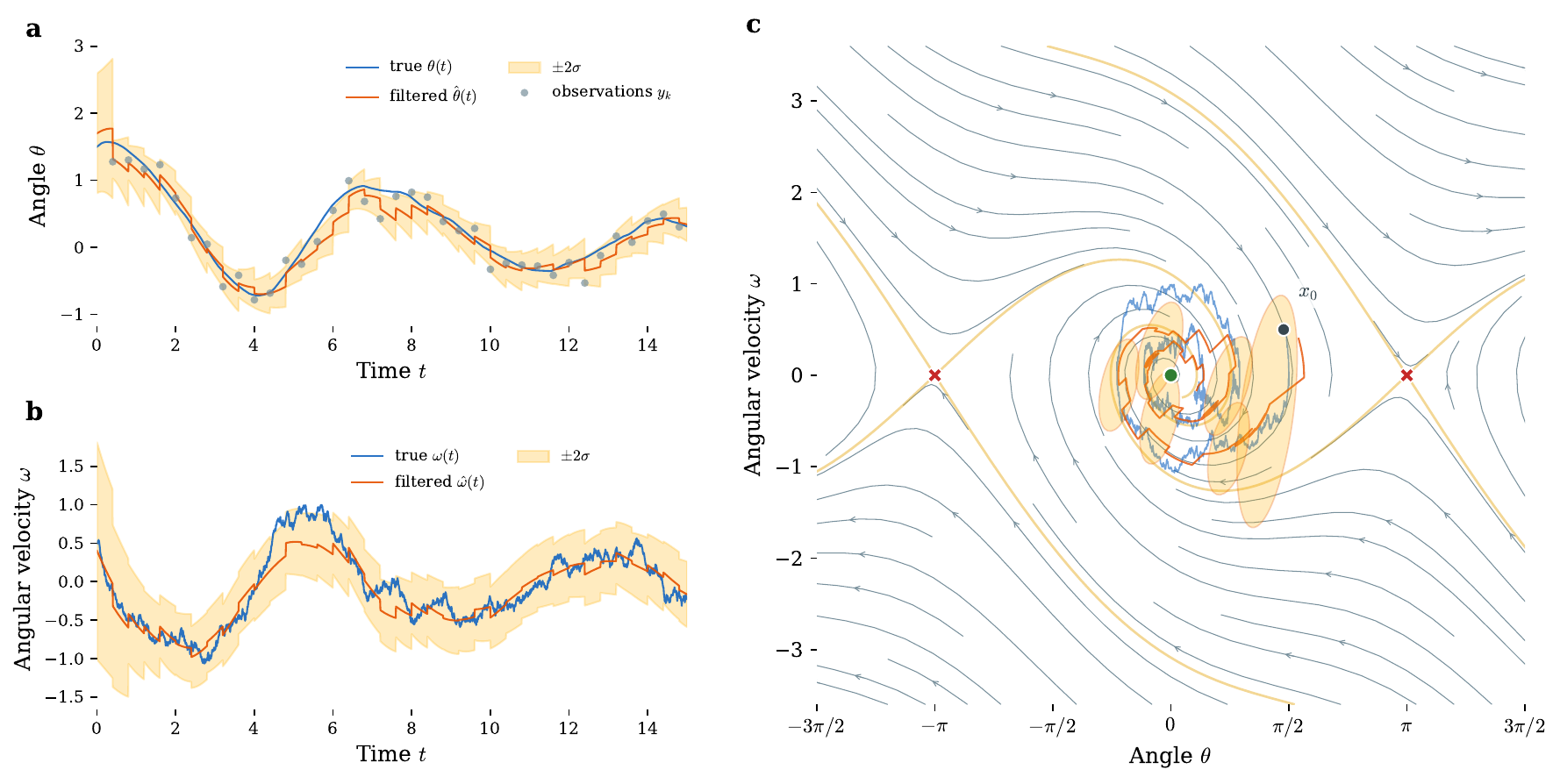}
    \caption{\textbf{(a)} Time series of the angle: \textcolor{penTrue}{true $\theta(t)$} (\textcolor{penTrue}{---}), discrete noisy \textcolor{penObs}{observations} $y_k$ (\textcolor{penObs}{$\bullet$}), and the \textcolor{penFilt}{filtered estimate $\hat{\theta}(t)$} (\textcolor{penFilt}{---}) with its $\pm 2\sigma$ band. The band widens between observations (prediction step) and narrows when a measurement arrives (update step). \textbf{(b)} Time series of the angular velocity: no observations are available, but the \textcolor{penFilt}{filtered estimate $\hat{\omega}(t)$} (\textcolor{penFilt}{---}) tracks the \textcolor{penTrue}{true $\omega(t)$} (\textcolor{penTrue}{---}) by exploiting the relationship $\dot\theta=\omega$, with a visibly wider $\pm 2\sigma$ band. \textbf{(c)} Posterior covariance ellipses in the phase plane along the filtered trajectory (\textcolor{penFilt}{---}), starting from $x_0$ (\textcolor{penIC}{$\bullet$}). The ellipses are narrow along $\theta$ (directly observed) and elongated along $\omega$ (inferred indirectly), reflecting the asymmetry induced by the observation function $h$.
}
    \label{fig:4}
\end{figure*}

%
\circledlabel{S3} \textbf{Partial observations.} Suppose we observe only the angle of the pendulum, corrupted by Gaussian noise:
\begin{equation}
y_k = \theta(t_k) + v_k, \quad v_k \sim \mc{N}(0, \sigma^2_{v}).  
\end{equation}
The angular velocity is not measured directly. A filter must infer $\omega$ from the way $\theta$ changes across successive observations. Since $\dot{\theta} = \omega$, a rapidly changing sequence of angle measurements implies a large velocity. The resulting uncertainty has a characteristic shape in the phase plane: the posterior covariance $P_k$ forms an ellipse that is narrow along $\theta$ (directly observed, constrained up to the noise level $\sigma_{v}$), and elongated along $\omega$ (never observed directly, inferred only through the dynamical relationship $\dot{\theta} = \omega$).
\clearpage 

\begin{figure*}[!h]
    \centering
    \includegraphics[width=\linewidth]{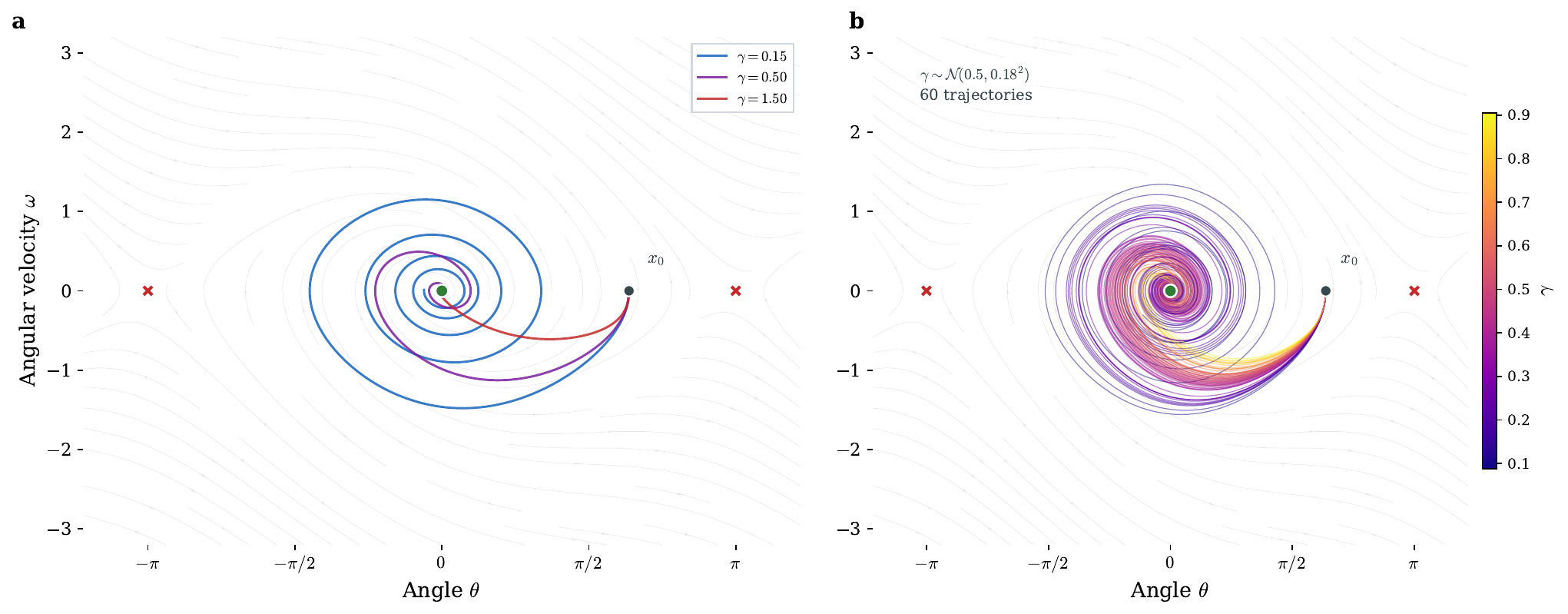}
    \caption{\textbf{(a)} Three trajectories from the same initial condition $x_0$ (\textcolor{penIC}{$\bullet$}) under different values of the damping coefficient: \textcolor{penGammaLow}{$\gamma=0.15$} (\textcolor{penGammaLow}{---}), \textcolor{penGammaMid}{$\gamma=0.50$} (\textcolor{penGammaMid}{---}), and \textcolor{penGammaHigh}{$\gamma=1.50$} (\textcolor{penGammaHigh}{---}). The same system produces qualitatively different paths to the \textcolor{penStable}{stable focus} (\textcolor{penStable}{$\bullet$}), from many oscillations under light damping to a near-monotonic approach under heavy damping. \textbf{(b)} An ensemble of 60 trajectories from the same $x_0$ with $\gamma \sim \mathcal{N}(0.5,0.18^2)$ truncated to $(0,\infty)$, colored by the sampled value of $\gamma$. Trajectories are initially close but progressively fan out over time, since the cumulative effect of different damping rates accumulates along the flow.}
    \label{fig:5}
\end{figure*}

\circledlabel{S4} \textbf{Parameter uncertainty.} Consider the pendulum \eqref{eq:pend} with uncertain damping $\gamma$. Suppose our posterior $p(\gamma \,|\, \mc{D})$ is concentrated around $\gamma = 0.5$ but with non-trivial spread. For each sampled value of $\gamma$, the phase portrait changes, since the rate at which trajectories spiral inward varies and the shape of the separatrix changes. An ensemble of trajectories, each evolved under a different $\gamma$ drawn from the posterior, fans out over time. Initially, the trajectories remain close because their short-term behavior is similar, but they progressively diverge as differences in damping accumulate. 
Unlike \circledlabel{S1}, where an ensemble near the stable focus contracts under a fixed flow, parameter uncertainty can produce substantial transient fan-out because different parameter values generate different flows. In this damped example with only $\gamma$ uncertain, the ensemble ultimately reconverges to the same stable equilibrium, although the rate and path depend on $\gamma$. More generally, parameter uncertainty can persist asymptotically when it changes the attractor, basin structure, oscillation frequency, or qualitative regime.

\clearpage 

\begin{figure*}[!h]
    \centering
    \includegraphics[width=\linewidth]{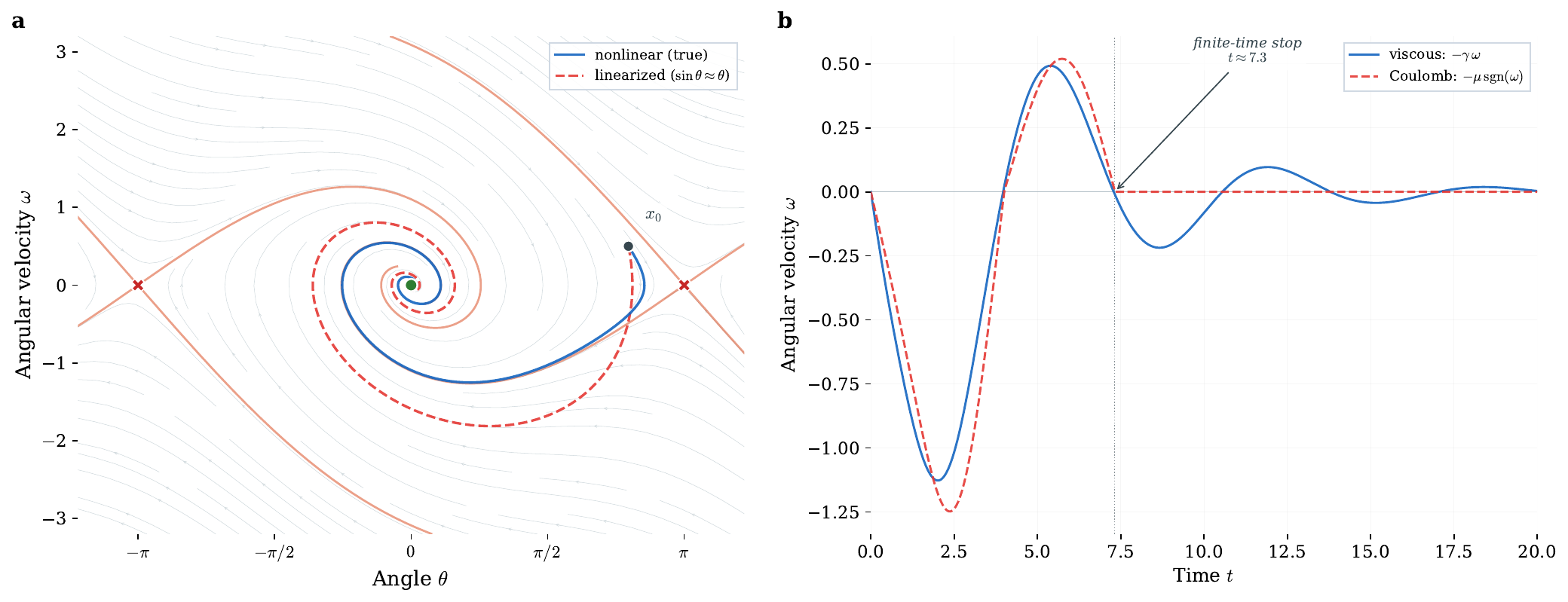}
    \caption{\textbf{(a)} Same initial condition $x_0$ (\textcolor{penIC}{$\bullet$}) evolved under two different models: the \textcolor{penNL}{nonlinear} pendulum (\textcolor{penNL}{---}) and its \textcolor{penLin}{linearized} version $\sin\theta \approx \theta$ (\textcolor{penLin}{-- --}). Near the origin the two models agree, but at large amplitudes, where the trajectory approaches the \textcolor{penSeparatrix}{separatrix} (\textcolor{penSeparatrix}{---}), they diverge: the linearization has no separatrix and no notion of full rotation, so it takes a structurally different path through phase space. \textbf{(b)} Angular velocity over time under two friction models, \textcolor{penNL}{viscous $-\gamma\omega$} (\textcolor{penNL}{---}) and \textcolor{penLin}{Coulomb $-\mu\,\mathrm{sgn}(\omega)$} (\textcolor{penLin}{-- --}), starting from the same initial condition. The two trajectories are nearly indistinguishable for most of the motion, but they diverge qualitatively near rest; the Coulomb model exhibits a \emph{finite-time stop} at $t \approx 7.3$, while the viscous model only approaches zero asymptotically. 
}
    \label{fig:6}
\end{figure*}
\circledlabel{S5} \textbf{Structural uncertainty.} Consider the pendulum \eqref{eq:pend}. For small angles, $\sin \theta \approx \theta$, and the dynamics are well-approximated by the linear system 
\begin{equation} \label{eq:pend-lin}
    \ddot{\theta} + \gamma \dot{\theta} + \frac{g}{\ell} \theta = 0.
\end{equation}
The phase portrait is qualitatively similar to that of the true nonlinear pendulum near the origin. Trajectories spiral inward toward a stable focus, and the eigenvalues of the linearized system match those of the nonlinear system at equilibrium. For trajectories that remain close to the equilibrium, the linear model is an excellent approximation, and no amount of data from such trajectories would suggest that anything is missing. Globally, however, the picture changes completely. The linear model has no separatrix, no saddle points, and no qualitative distinction between oscillation and rotation. In particular, it cannot represent the pendulum going over the top as a rotation on the phase cylinder. A trajectory of the linearized system that starts with high velocity does not correspond to such a rotation. Instead, in the linear tangent-plane model it remains a damped oscillator; after a possibly large excursion in $\theta$, it spirals inward toward the origin. Thus, the linearization breaks down precisely in regions where the true system's global phase-space structure is essential, including separatrices, saddle points, rotations, and the periodic identification of $\theta$. No filtering, no parameter tuning, and no additional data can make this linear model represent that missing global structure. The model is structurally incapable of representing the global dynamics of the pendulum, regardless of how accurately its parameters are estimated.

A different kind of structural error arises when the damping mechanism itself is misspecified. Suppose, for example, that someone proposes the true damping to be not viscous, $-\gamma \omega$, but Coulomb friction, $-\mu \, \rm{sgn}(\omega)$. The two models differ qualitatively. Under Coulomb friction, the pendulum comes to exact rest in finite time and then remains there. Under viscous damping, by contrast, the approach to rest is asymptotic. For trajectories that do not probe the small-velocity regime, the two models may appear nearly indistinguishable; only careful observation near rest reveals which description is correct. Choosing the wrong friction law is therefore a structural error that data alone may not readily expose.


\end{document}